\documentclass[letterpaper]{article}
\usepackage[submission]{aaai23}
\usepackage{times}
\usepackage{helvet}
\usepackage{courier}
\usepackage[hyphens]{url}
\usepackage{graphicx}
\usepackage{natbib}
\usepackage{caption}
\usepackage{algorithm}
\usepackage{algorithmic}
\usepackage{tabularx}
\usepackage{multirow}

\newtheorem{definition}{Definition}

\usepackage{newfloat}
\usepackage{listings}
\DeclareCaptionStyle{ruled}{labelfont=normalfont,labelsep=colon,strut=off}
\floatstyle{ruled}
\newfloat{listing}{tb}{lst}{}
\floatname{listing}{Listing}

\title{On the Applicability of Synthetic Data for Re-Identification

in Warehousing Logistics\\
}
\author{
    Written by AAAI Press Staff\textsuperscript{\rm 1}\thanks{With help from the AAAI Publications Committee.}\\
    AAAI Style Contributions by Pater Patel Schneider,
    Sunil Issar,\\
    J. Scott Penberthy,
    George Ferguson,
    Hans Guesgen,
    Francisco Cruz\equalcontrib,
    Marc Pujol-Gonzalez\equalcontrib
}
\title{On the Applicability of Synthetic Data for Re-Identification

in Warehousing Logistics}
\author {
    Jérôme Rutinowski\textsuperscript{\rm 1},
    Bhargav Vankayalapati\textsuperscript{\rm 1},
    Nils Schwenzfeier\textsuperscript{\rm 2},\\
    Maribel Acosta\textsuperscript{\rm 3},
    Christopher Reining\textsuperscript{\rm 1}
}
\affiliations {
    \textsuperscript{\rm 1} TU Dortmund University, Germany: \{jerome.rutinowski; bhargav.vankayalapati; christopher.reining\}@tu-dortmund.de\\
    \textsuperscript{\rm 2} University of Duisburg-Essen, Germany: nils.schwenzfeier@uni-due.de\\
    \textsuperscript{\rm 3} Ruhr-University Bochum, Germany: maribel.acosta@ruhr-uni-bochum.de
}

\usepackage{bibentry}

\begin{document}

\maketitle

\begin{abstract}
This contribution demonstrates the feasibility of applying Generative Adversarial Networks (GANs) on images of EPAL pallet blocks for dataset enhancement in the context of re-identification.
For many industrial applications of re-identification methods, datasets of sufficient volume would otherwise be unattainable in non-laboratory settings.
Using a state-of-the-art GAN architecture, namely CycleGAN, images of pallet blocks rotated to their left-hand side were generated from images of visually centered pallet blocks, based on images of rotated pallet blocks that were recorded as part of a previously recorded and published dataset.
In this process, the unique chipwood pattern of the pallet block surface structure was retained, only changing the orientation of the pallet block itself.
By doing so, synthetic data for re-identification testing and training purposes was generated, in a manner that is distinct from ordinary data augmentation.
In total, 1,004 new images of pallet blocks were generated.
The quality of the generated images was gauged using a perspective classifier that was trained on the original images and then applied to the synthetic ones, comparing the accuracy between the two sets of images.
The classification accuracy was 98\% for the original images and 92\% for the synthetic images.
In addition, the generated images were also used in a re-identification task, in order to re-identify original images based on synthetic ones.
The accuracy in this scenario was up to 88\% for synthetic images, compared to 96\% for original images.
Through this evaluation, it is established, whether or not a generated pallet block image closely resembles its original counterpart.
\end{abstract}

\section{Introduction}
Generative Adversarial Networks (GANs) have shown to be an efficient way to generate synthetic data (herein defined as data generated by GANs or similar generative methods, which typically are supposed to resemble original data in at least one metric), producing better results than other generative methods \cite{cai2021generative,goodfellow2016nips,wang2021generative}.
The generation of such data can be helpful for data-driven approaches and in situations in which the acquisition of sufficient original data is not feasible.
This, for example, is the case when trying to examine the scalability of a solution which retrieves its input from a high volume database, for which the gathering of enough original data would not be possible in an economically feasible and timely manner.

An example of one such case is the development of a re-identification solution for warehousing entities, as presented in \cite{rutinowski2021}.
In this work, the authors focused on applying the concept of pedestrian re-identification to chipwood pallet blocks. 
Re-identification is commonly defined as the identification of a previously recorded subject over a network of cameras \cite{Mang2021} and assigning a unique descriptor to it instead of assigning it to a class holding multiple entities of the kind.  
The motivation of this contribution lies in the absence of an inbuilt identification feature on Euro-pallets, which would be of great value for the automation of warehousing processes.
Thus far, the standardized Euro-pallet, whose specifications are defined in a European Norm \cite{din_standards_committee_packaging_pallet_2004}, guaranteeing production standards and interoperability,  rely on methods that use artificial features (e.g., barcodes, QR-codes or RFID).
However, equipping pallets with these artificial features implies further expenses, labor, and the risk of illegibility after an elongated period of use, due to material deterioration.
Even though it has been demonstrated that the re-identification of pallet blocks can be achieved in a laboratory environment \cite{rutinowski2021}, it is not evident that the system that has been developed would yield similar results (i.e., a similar re-identification accuracy) if the dataset were of much greater extent.
It is not evident either, whether real-time performance (i.e., registering a pallet and receiving a re-identification result in a matter of a few seconds) could still be achieved.
These two aspects however do matter, since hundreds of millions of Euro-pallets are in constant circulation in the industry \cite{deviatkin2020carbon}.
Even so, the acquisition of millions of images of pallet blocks is not feasible in a timely manner, which is why the use of synthetic data is of such great benefit in this very case.

Therefore, this contribution presents a GAN based approach for the generation of synthetic pallet block images.
The result of this approach is that the centered input images of pallet blocks from the dataset \emph{pallet-block-502} \cite{rutinowski_pallet-block-502_2021} are rotated to their left-hand side, while maintaining the unique chipwood pattern of their surface structure. 
By doing so, a new image of the same pallet block from another perspective is generated.
This dataset enhancement approach is distinct from ordinary data augmentation, in which an image would be flipped, rotated, etc.
Since to the best of our knowledge, the synthetic generation of pallet block images is a novel application of GANs, no standard practices for performance or result quantification have been encountered.
As to assess the quality of the synthetic images, a classifier is therefore trained on original centered and rotated images and applied to a hold-out set of original and synthetic images.
The results of applying this classifier on the two sets of images are subsequently compared, determining whether the synthetic images are classified accordingly.
In addition, the synthetic images are used in the re-identification pipeline presented in \cite{rutinowski2021}, as to determine whether the retention of the chipwood pattern during the rotation of the pallet blocks was successful, i.e., whether the same pallet block, only rotated instead of centered, would still get assigned the same ID and thus be recognized accordingly in the re-identification scenario. 
Therefore, while the main contribution of this work is the generation of synthetic data, the classification and re-identification tasks are used as benchmarks to validate the applicability of said data.

The remainder of this work is structured as follows: 
Section~\ref{sec:related_work} discusses the relevant research related to the subject of GANs and their applications as well as the concept of deep learning based re-identification.
Section~\ref{sec:approach} lays out the approach and methodology proposed in this work. 
Section~\ref{sec:results} presents the evaluation of the results that have been obtained.
Finally, Section~\ref{sec:conclusion} discusses these results and provide some suggestions on the research that could further be conducted based on this contribution.

\section{Related Work}
\label{sec:related_work}
The relevant state of the art for this contribution comprises GANs for dataset enhancement and deep learning based re-identification.
The two will be discussed briefly in this Section.

\subsection{Generative Adversarial Networks}
Generative Adversarial Networks can be understood as a two-player zero-sum game, in which (at least) two neural networks compete against each other in the form of a minimax optimization problem \cite{wang2017generative}. 
On the most rudimentary level, one of these networks, called the generator, tries to generate new, synthetic data from a set of original data or from noise that it is given as an input \cite{wang2021generative}.
The discriminator then acts as the generator's adversary, often in the form of a binary classifier, trying to distinguish between original and synthetic data \cite{wang2021generative}.
Starting from this basic premise \cite{goodfellow2016nips}, different kinds of GAN architectures and approaches have since been developed.

One such development is pix2pix GAN \cite{pix2pix}, which is a conditional GAN, meaning that a conditional variable (i.e., a label) is added to both the generator and the discriminator model during training \cite{Pan19survey}.
By adding this additional information during training, the data generation process is affected and can therefore effectively be influenced by altering the conditional variable \cite{langr2019gans}.
Isola et al. argue that many image processing tasks revolve around a translation from input to output, which can be abstracted as mapping pixels to pixels \cite{pix2pix}.
The authors claim that this challenge can therefore be solved by conditional GANs that only need the respective training data but keep their same architecture and objective (i.e., loss function) for each task.
To prove this claim, Isola et al. used pix2pix GAN for various image generation and semantic segmentation tasks, such as generating photo-realistic images from sketches or converting daytime images to nighttime images.
Pix2pix GAN was later proceeded by pix2pixHD \cite{pix2pixHD}, which provides photorealistic image-to-image translation at resolutions of up to $2048$ × $1024$ px.
Like its predecessor, pix2pixHD translates semantic label maps into images.

\begin{figure}[htbp]
\centerline{\includegraphics[width=1\columnwidth]{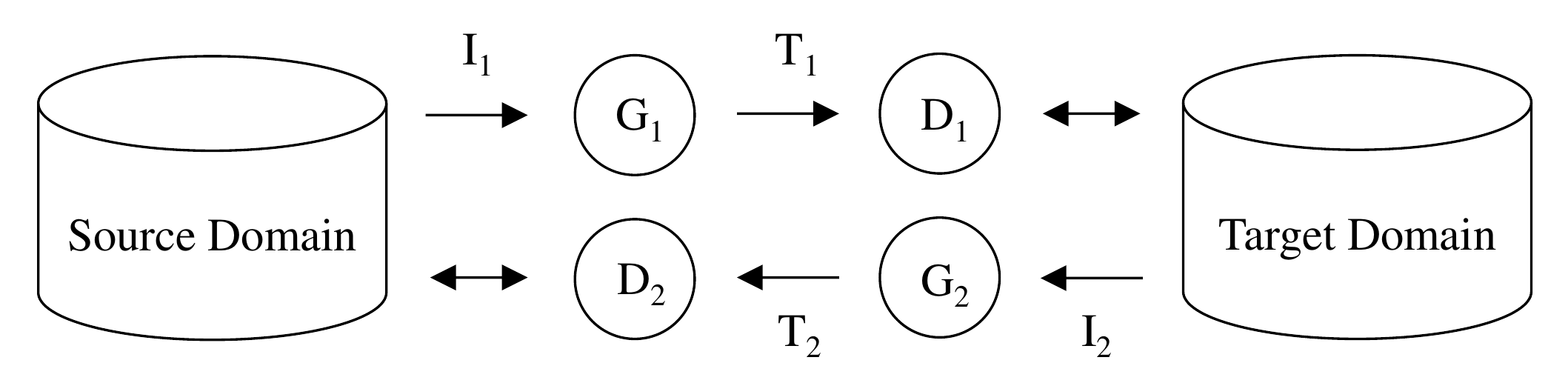}}
\caption{A schematic depiction of the concept of cycle consistency, as described in \cite{CycleGAN2017}.}
\label{figure1}
\end{figure}

Another GAN architecture that is of interest is CycleGAN \cite{CycleGAN2017}.
This GAN has a deep convolutional architecture that learns a mapping between one image domain and another, using unsupervised learning to train two generators and discriminators.
While the previously mentioned pix2pix GAN uses an explicit pairing between labeled inputs and outputs, such paired data can be hard to come by in real life applications.
In such cases, an unsupervised approach is advantageous, in the specific case of CycleGAN the use of cycle consistency.
Cycle consistency means that one generator $G_1$ translates input $I_1$ from the source domain to the target domain, where a discriminator $D_1$ tests whether the translated input $T_1$ is distinguishable from target domain samples \cite{CycleGAN2017}.
Simultaneously, another generator $G_2$ translates $I_2$ from the target domain to the source domain, where another discriminator $D_2$ again tests whether the translated input $T_2$ is distinguishable from source domain samples (see Figure \ref{figure1}).
Due to this functionality, even if no precisely matched training image pairings are available, CycleGAN can generate synthetic images from the target domain based on images from the source domain.

\subsection{Deep Learning based Re-Identification}
\label{sec:DL_based_Re-ID}
As has been hinted in the introduction of this contribution, for the purpose of re-identification, the subject of interest is assigned a descriptor when first recorded by one of the cameras in the camera network.
This descriptor assignment, which is similar to a class assignment, as would be common in classification tasks, is what the re-identification of a previously recorded subject at a later point in time is based on.
The difference, compared to an ordinary classification task, lies in the uniqueness of the descriptor, meaning that classes containing only a single instance are created (when re-identification is treated as a classification task and not a metric learning task, which would be a valid alternative \cite{metric1,metric2}).

Re-identification is commonly used in the context of pedestrian surveillance and in this context commonly realized by use of deep learning based methods \cite{Mang2021,metric1,zheng2016person}.
Apart from pedestrians, other entities, such as vehicles, have been the subject of re-identification \cite{Rong2021AVR,wei2018}, but the focus remains strongly on humans.
In a previous publication \cite{rutinowski2021}, it was demonstrated that re-identification methods can also be of value in the context of warehousing logistics.
For this purpose, a state-of-the-art pedestrian re-identification framework, Torchreid \cite{torchreid}, was used in conjunction with PCB\_P4 \cite{sun2018beyond} on a self developed dataset called pallet-block-502 \cite{rutinowski_pallet-block-502_2021}.
This dataset contains $10$ images each of $502$ pallet blocks of Euro-pallets made out of chipwood.
The $10$ images are made up of five different perspectives (central perspective, left/right-hand side rotation, left/right-hand side shift) and two lighting modes (natural and artificial lighting).
With this setup, it was possible to re-identify the respective pallet blocks based on their individual, unique chipwood pattern with an accuracy of at least 96\%, depending on the matching scenario.

\section{Methodological Approach}
\label{sec:approach}
In this work, we use a subset of the dataset pallet-block-502 \cite{rutinowski_pallet-block-502_2021}, presented in Section~\ref{sec:DL_based_Re-ID}. 
It is a suitable dataset for the task at hand, since it contains multiple labeled images of pallet blocks, taken from different perspectives, with different lighting conditions.
We focus on images belonging to two distinct perspectives, which will be our source and target domains, namely the central (C) and left-hand side rotation (RL) perspectives. 
Formally, the input data can be defined as follows: 

\begin{definition}
Let $\mathcal{D}$ be a dataset composed of images of $N$ pallet blocks from a central $\mathcal{C}$ and left-hand side rotation $\mathcal{RL}$ perspective, s.t. $\mathcal{C} \cap \mathcal{RL}=\emptyset$. 
In $\mathcal{D}$, there are two distinct images for each perspective ($\mathcal{C}$ and $\mathcal{RL}$) of a single pallet block $i$ ($1 \leq i \leq N$) with varying lighting conditions.
\end{definition}

In this work, the problem is defined as given an image $c_i \in \mathcal{C}$ of a pallet block $i$, generate a synthetic image  for the RL perspective. 
We denote these synthetic images as $\hat{r}_i$ and their set as $\widehat{\mathcal{RL}}$. 
To address this problem, in Section~\ref{sec:gan} we first present the GAN architecture that was chosen for these experiments and discuss the way in which it was applied to our dataset.
Then, in Section~\ref{sec:tasks}, we present the downstream tasks that are used to assess the quality of the synthetic images that were generated by the GAN.

\subsection{Learning Process: GAN Selection \& Training}
\label{sec:gan}
\begin{figure}[thbp]
\centerline{\includegraphics[width=1\columnwidth]{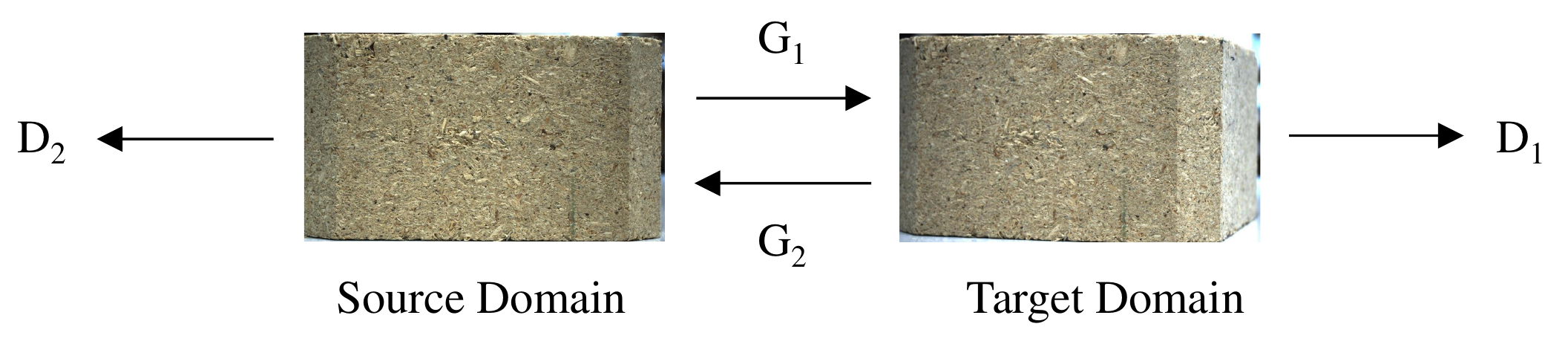}}
\caption{Examples of centered (source domain) and rotated (target domain) images used for CycleGAN training (ID $187$).}
\label{figure2}
\end{figure}

Based on the C and RL perspective subset, consisting of two times $1,004$ images of the dataset, we aim to again generate $1,004$ images from the RL perspective, by providing $1,004$ corresponding images from the C perspective, while preserving the chipwood surface structure of the respective pallet block. 
Figure \ref{figure2} shows an example of a single pallet block from these two rotational domains, embedded in the conceptualized representation of an adversarial network with cycle consistency.

With this aim in mind, image-to-image translation can be used to create RL perspective images from C perspective images.
While pix2pix GAN could be used to create a supervised one-to-one mapping between the two domains, it requires a paired set of C and RL images.
This, however, cannot be guaranteed for future use cases, reducing the appeal of pix2pix GAN for our intents and purposes.
CycleGAN, on the other hand, uses unsupervised learning to train two generators and two discriminators.
By doing so, even without precisely matched training image pairings, it can generate synthetic images from the RL domain based on input images from the C domain, which is the perspective shift we will focus on.
For this reason, we will use CycleGAN for our experiments.

Therefore, the CycleGAN generator in this scenario consists of an encoder and a decoder, with the encoder being a ResNet \cite{he2016deep}  with convolutional layers that map the input (i.e., a C perspective image) to a small feature representation and the decoder being a ResNet with transpose convolutional layers, transforming the feature representation it obtains into a transformed output (i.e., an RL perspective image).
In this unsupervised setting, a single generator may map multiple input images to a single output, leading to what is called mode collapse \cite{lala2018evaluation}, which means that outputs with little to no variance would be generated.
To avoid this pitfall, CycleGAN uses cycle consistency, in this case therefore inversely performing RL to C transformation while simultaneously training another generator to perform C to RL transformation.
Hence, cycle consistency allows CycleGAN to correctly map C perspective images to RL perspective images, while simultaneously preserving the features of the input image in the generated image.

CycleGAN's two discriminator networks work analogously to its generators.
The discriminator network is a convolutional network that distinguishes between original and synthetic images, classifying them accordingly.
During each iteration the discriminator is applied to the current batch of synthetic images generated by the respective generator as well as $50$ images generated during previous iterations.
This enables the discriminator to generalize well in the target domain, i.e., RL perspective images.
CycleGAN optimizes the adversarial loss and the cycle consistency loss, which quantifies the difference between the original C perspective input image, an image translated into the RL perspective and back again, to produce what it considers to be appropriate output images.
Additionally, an identity loss is taken into account as well, in order to retain the image's color space.
As a result, even when using an unpaired dataset, CycleGAN can generate RL perspective images from a C perspective input image.

\subsection{Formulating the Evaluation Tasks}
\label{sec:tasks}
Different GAN evaluation settings do exist \cite{borji2019pros,borji2022pros} and can in many ways be valuable tools that enable researchers to assess the performance of their GAN.
However, these evaluations  primarily focus on GAN training, leaving room for ambiguity when it comes to the application of a trained generator on new input data.
This however, is a key factor in many use cases, such as our own, which determines the usefulness of a trained generator.
Given this background, we developed our own evaluation method, tailored to our use case.
Therefore, as to evaluate the performance of the images generated by the trained CycleGAN model, the two following evaluation tasks are performed:

\subsubsection{Image perspective classification}
In this task, the class is defined as the perspective of the image. This is formally defined as follows:

\begin{definition}
Given an image $x$, the image perspective classification task is defined as the problem of determining if $x \in \mathcal{C}$ or $x \in \mathcal{RL}$. 
\end{definition}

In this work, a classifier is trained on $80\%$ of the C and RL pallet block images of the dataset pallet-block-502, with the aim of distinguishing between these two types of perspectives ($1,608$ images in total, $804$ per class).
The trained model is then applied to the hold-out dataset consisting of the remaining $20\%$ of the respective dataset images ($400$ images in total, $200$ images per class), thereby establishing a benchmark for how accurately an original yet unknown image can be classified by its perspective.
Finally, the model is applied to a randomly chosen subset (of the same size as the hold-out dataset) of the synthetic images and the classification accuracy on both sets of images compared to one another.

\subsubsection{Re-identification of pallet blocks based on synthetic images} Based on our definition of the input data, the task of re-identification is  defined as: 

\begin{definition}
Given a query set $\mathcal{Q} \subset \mathcal{D}$ and a gallery set $\mathcal{G} \subset \mathcal{D}$ with $\mathcal{Q}~\cap~\mathcal{G} = \emptyset$, the problem of re-identification of an image $x_i \in \mathcal{Q}$ of pallet block $i$ is to find an image $y_j \in \mathcal{G}$ of pallet block $j$ such that $i=j$.
\end{definition}

The re-identification pipeline from \cite{rutinowski2021} is used to match an original image of a pallet block to a synthetic image of the same pallet block (original C to synthetic RL perspectives and original RL to synthetic RL perspectives).
The accuracy, with which the re-identification could be performed is then compared to the results for the same scenario, that are obtained using only original images, matching the C and RL perspectives.

The dimensions of the generator's input and output images are set to $1280$ × $640$ px and therefore a $2$:$1$ aspect ratio, while the images from pallet-block-502 have an aspect ratio of approximately $1.7$:$1$, due to the images being cropped automatically by YOLO.
Taking this aspect ratio discrepancy into account, two modes of re-identification result evaluation will be carried out.
The first one will apply the exact approach used in \cite{rutinowski2021}.
The second one (subsequently called modified re-identification) will apply projective transformations and Gaussian blur to the training images and will center-crop the generated images to an aspect ratio of 1.7, as to match the aspect ratio of pallet-block-502.

\section{Experimental Evaluation}
\label{sec:results}
Before displaying and analyzing the results of this contribution, the experimental configuration, in terms of software and hardware usage, will be laid out.

\subsection{Experimental Configuration}
\setcounter{footnote}{0}
For the experiments in this contribution, three different networks were used.
First of all, we used a Tensorflow implementation of CycleGAN\footnote{https://github.com/LynnHo/CycleGAN-Tensorflow-2}.
The generators and discriminators of this network were trained for $200$ epochs, with nine residual blocks for the generator architectures, using the Adam optimizer with a learning rate of $0.0002$ and the corresponding $\beta_1$ argument set to 0.5.
These and all unmentioned parameters are the recommended default settings provided by the authors of \cite{CycleGAN2017}, taking the selected resolution into account.
Next, we trained a perspective classifier, as described in Section~\ref{sec:tasks}, again using the Tensorflow library.
It employs Adam as an optimizer (with the default learning rate of $0.001$), sparse categorical cross entropy as a loss function and was trained for $20$ epochs.
The architecture and further hyperparameters of the classifier can be seen in Table~\ref{classifier_table}.

\bgroup
\def\arraystretch{1.2}
\begin{table}[htbp]
\centering

\setlength\tabcolsep{2.5pt}
\resizebox{\columnwidth}{!}{
\begin{tabular}{lrl}
\hline
Layer     & \multicolumn{1}{c}{Shape}          & \multicolumn{1}{c}{Hyperparameters} \\ \hline
Input     & 640 x 1280 x 3 & - \\
Conv2D    & 640 x 1280 x 16                    & Filter size: 16; Stride: 3; Activation: ReLU \\
MaxPool2D & 320 x 640 x 16                     & Stride: 2                                                                                 \\
Conv2D    & 320 x 640 x 32                     & Filter size: 32; Stride: 3; Activation: ReLU\\
MaxPool2D & 160 x 320 x 32                     & Stride: 2                                                                                 \\
Conv2D    & 160 x 320 x 64                     & Filter size: 64; Stride: 3; Activation: ReLU \\
MaxPool2D & 80 x 160 x 64                      & Stride: 2                                                                                 \\
Flatten   & 819200                             & -                                                                                         \\
Dense     & 128                                & Activation: ReLU                                                                          \\
Dense     & 2                                  & Activation: Linear                                                                              \\ \hline
\end{tabular}}
\caption{Architecture of the pallet block perspective classifier.}
\label{classifier_table}
\end{table}
\egroup

Finally our re-identification method, which is described in detail in \cite{rutinowski2021}, was applied to the data in its modified and unmodified manner (see Section~\ref{sec:tasks}).
For the image perspective classification task, accuracy (defined as the amount of correctly classified images divided by the total number of all images) was the evaluation metric of choice.
Additionally, ranked accuracy (from rank 1 to rank 5) was used for the re-identification task, since unlike the herein described perspective classification, re-identification is not a binary classification task.

The experiments and evaluations described throughout this work   were run on an NVIDIA DGX-2 (equipped with $16$ NVIDIA Tesla V100 GPUs and a 24-core Intel Xeon Platinum 8168 CPU).
To support the reproducibility of our results, 
the code used for this work and the resulting images can be found online\footnote{https://github.com/jerome-rutinowski/gan\_data\_for\_re-id}.%}.

\subsection{Synthetic Data Generation Results}
\begin{figure}[ht]
\centerline{\includegraphics[width=0.9\columnwidth]{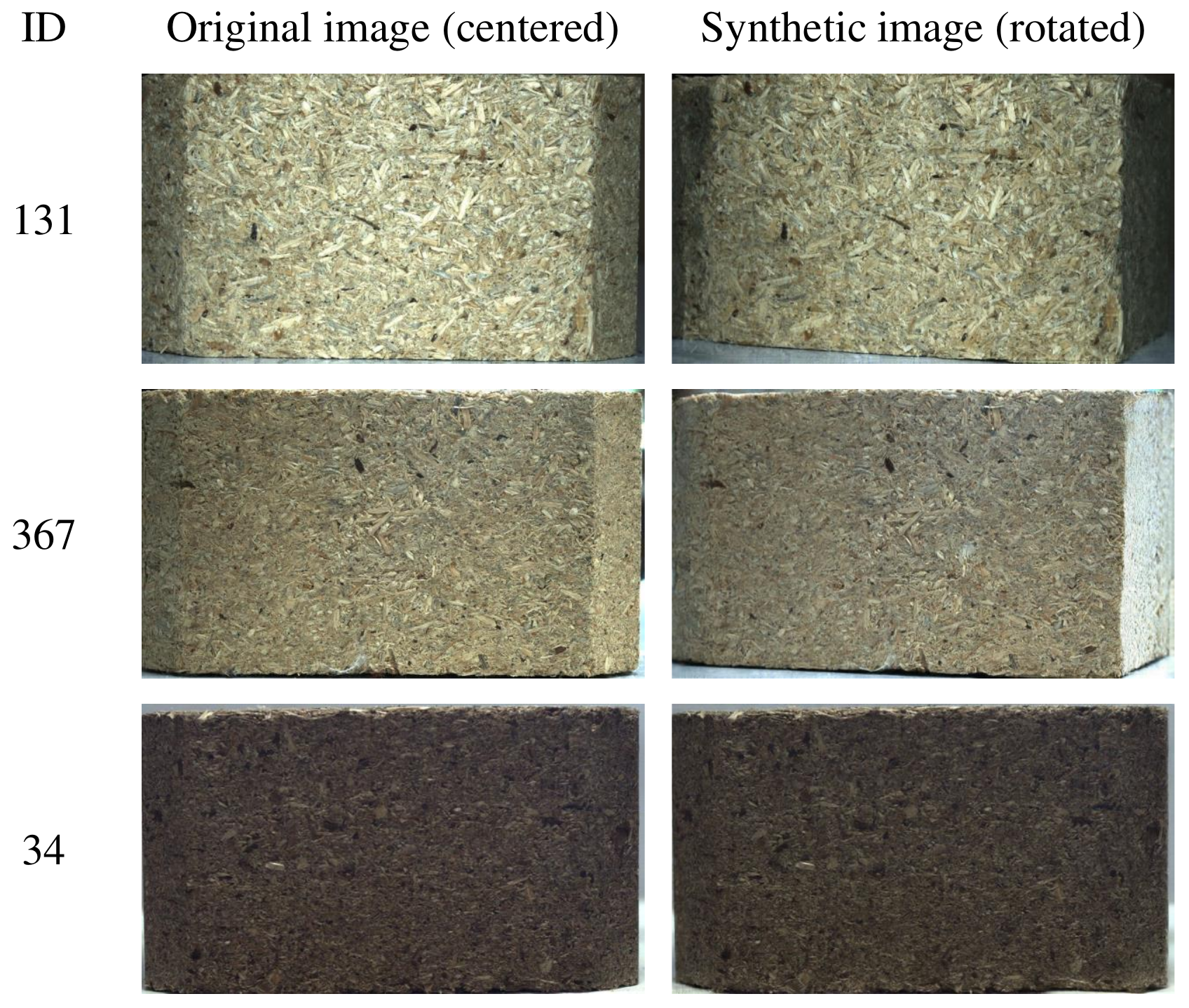}}
\caption{Original and corresponding synthetic images generated by applying a left-hand side rotation to a centered input image.}
\label{figure3}
\end{figure}

By using the methods described in the previous Section, $1,004$ images of RL perspective pallet blocks were generated from their C perspective counterpart that they are based on, retaining their chipwood surface structure while only changing the perspective.
Some examples of generated images and their original counterparts can be seen in Figure~\ref{figure3}.

For original images that had a low degree of luminosity, as can be seen with ID 34 in Figure~\ref{figure3}, the generator was consistently unable to change their perspective, generating only a (visually) very similar copy of the original image.
This was the case for $102$ of the $1,004$ images that were generated.
The generator likely experienced mode collapse in these instances and the affected images were therefore excluded from further evaluation.
For the remaining majority of the data however, no mode collapse could be observed and the resulting images look visually promising.

\subsection{Results of the Classification Task}
\label{sec:Results_Classification}
Following the data generation process, a classifier was trained to discriminate between RL and C perspective images.
The classifier was then applied to the hold-out dataset of original images as well as a set of generated images of the same size ($200$ images each).
Even though the resulting images (see Figure~\ref{figure3}) seem very promising visually, a discrepancy can still be measured in terms of the perspective classification accuracy.
While the original images are being classified with an accuracy of $98\%$, their synthetic counterparts score an accuracy of $92\%$, meaning that the evaluation displayed a $6\%$ difference in classification accuracy.
This difference could be due to artifacts that might have been generated along with the new perspective of the image.
Additionally, the edges of the pallet blocks in the generated images, in some instances, are sharper and less natural looking than their original counterparts.
For the task of classification, a factor of uncertainty to also take into consideration is the vague definition of the terms “centered” and “rotated” during the creation of the original dataset.
During this process, no clear definition of these terms was given (i.e., in terms of an angle at which a pallet block ought to be facing the camera).
Since only a broad visual notion of the terms “centered” and “rotated” was used, the accuracy with which the dataset setup could be reproduced is reduced (i.e., it would be difficult to replicate the images, without knowing from what specific angle they should be taken) and categorical ambiguity is persistent.

\subsection{Results of the Re-Identification Task}
In addition to the classification task described in the Subsection above, the re-identification method from \cite{rutinowski2021} was applied to the original images, as described in Section~\ref{sec:tasks}.
As can bee seen in Table~\ref{results_table}, for the re-identification results, a distinct difference is to be noted between the results obtained using the exact methodology of \cite{rutinowski2021} and the modified version described in Section~\ref{sec:tasks} (discrepancies of $23$\% to $58$\%). 

\bgroup
\def\arraystretch{1.2}
\begin{table}[htbp]
\centering

\setlength\tabcolsep{3pt}
\resizebox{\columnwidth}{!}{
\begin{tabular}{lrrr}
\hline
\multicolumn{1}{l}{}                  & \multicolumn{2}{c}{Evaluation type}                                                      \\ \cline{2-3}
\multicolumn{1}{l}{Metric} & 
 Re-identification [\%]      & Re-identification (modified) [\%] \\ 
 \hline

$Acc_{C \rightarrow RL}$                                 & \multicolumn{1}{r}{73} & \multicolumn{1}{r}{96}             \\
$Acc_{C \rightarrow  \widehat{RL}}$                                  & 33                      & 88                                  \\
$Acc_{RL \rightarrow \widehat{RL}}$                                    & 20                      & 78                                  \\[0.5ex] \hline
\end{tabular}}
\caption{Re-identification rank-1-accuracy ($Acc$) on synthetic and original images.}
\label{results_table}
\end{table}
\egroup

In this case, the results suggest that using similar aspect ratios is paramount to a successful re-identification.
In addition, we assume that the use of Gaussian blur is advantageous in this case as well, blurring pixel level patterns, that would seem irrelevant to humans but might be mistakenly treated as a relevant feature by the re-identification algorithm.
However, when observing the results that were obtained using these modifications of compatibility, the discrepancy between the re-identification accuracy obtained using original images and synthetic images ranged between $8$\% and $18$\%.
In the former case, this represents a result that is very similar to the classification discrepancy that was described in Section~\ref{sec:Results_Classification}.
Of particular interest however, is the lower accuracy for $Acc_{RL \rightarrow \widehat{RL}}$ compared to $Acc_{C \rightarrow \widehat{RL}}$.
This could potentially mean that the synthetically generated RL perspective images still have a greater degree of similarity to the C perspective images that they are based on, than the RL perspective images that they are trying to replicate.

\begin{figure}[htbp]
\centerline{\includegraphics[width=1\columnwidth]{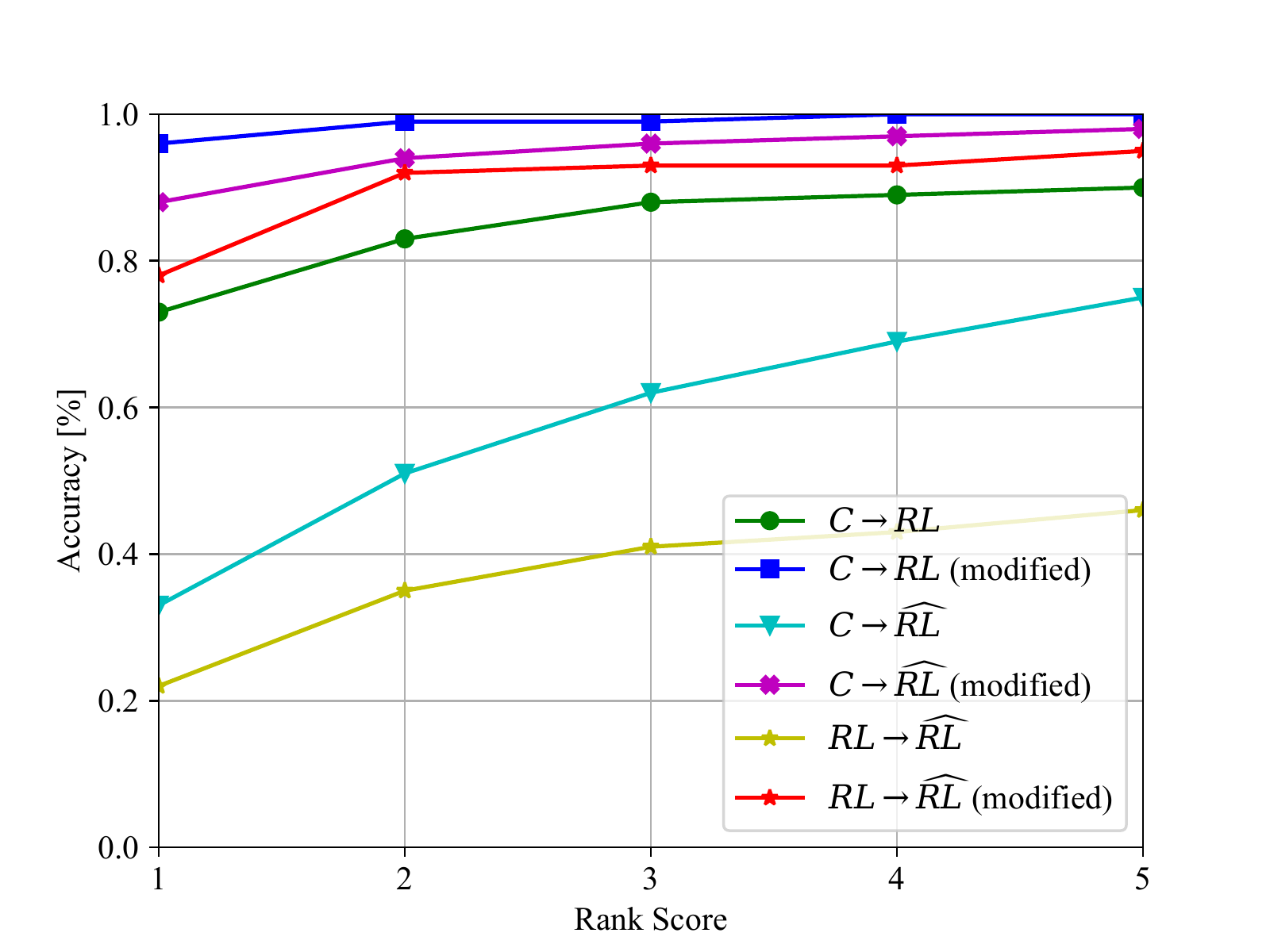}}
\caption{Ranked accuracy of the re-identification evaluation plotted as a CMC curve.}
\label{figure4}
\end{figure}

Finally, the re-identification results are shown in greater detail in Figure~\ref{figure4}, in the form of a CMC (Cumulative Match Characteristic) curve.
Taking more than the first rank accuracy into account, it can be seen that for all evaluation tasks using the modified re-identification method, results upwards of $90\%$ are consistently achieved.
For the un-modified re-identification method, an improvement in accuracy can be observed as well, as to be expected, with an increasing number of ranks.
However, until rank 5, a considerable difference in accuracy remains between the modified and the un-modified re-identification method.
This difference is pronounced enough, that $Acc_{C \rightarrow RL}$, which is performed only on original images, is consistently lower than $Acc_{C \rightarrow \widehat{RL}}$ (modified).

\section{Conclusion \& Future Work}
\label{sec:conclusion}
The results obtained in this contribution demonstrate that, in general, the generation of synthetic images of pallet blocks for the purpose of dataset enhancement is feasible.
Using the dataset pallet-block-502 and the state-of-the-art GAN architecture CycleGAN, $1,004$ images of rotated pallet blocks were generated from images of the same pallet blocks in a centered perspective.
For $102$ of these images, which were taken under comparatively low lighting conditions, no visual rotation could be perceived.
We assume this to be the case due to mode collapse, during the generator's training.
The remaining $902$ synthetic images closely resemble the original images in terms of their chipwood surface structure, but differ in the way the pallet blocks in the images are oriented towards the camera (i.e., they are now rotated to the left-hand side, instead of being centered).
Therefore, from a visual perspective, the aim of the herein described procedure was accomplished and new data could reliably be generated, enhancing the existing dataset.

Beyond visuals, the synthetic images were evaluated by using a classifier, trained on original images, that discriminates between rotated and centered perspectives.
Both original and synthetic images were run through the classifier and the classification accuracy was compared.
While the classification accuracy for original images was $98\%$, the classification accuracy for synthetic images was $92\%$, meaning that there remains a $6\%$ discrepancy.
This discrepancy implies that there is still a measurable difference between original and synthetic images, in terms of their perspective.
Additionally, the (modified) re-identification method presented in \cite{rutinowski2021} was applied to the synthetic images and the resulting re-identification accuracy of $88\%$ was compared to the accuracy of $96\%$ resulting from the use of original images only.
Again, the $8\%$ discrepancy shows that there remains a measurable difference between original and synthetic images, even in terms of their surface structure.
Therefore, while discrepancies still can be made out, we perceive the results obtained in this contribution as valuable and promising, further confirming the visually satisfactory results.

Finally, the work presented in this contribution can be improved, for instance, by using more input images for the GAN or by using a different GAN architecture to begin with.
Additionally, reducing the dependency on lighting conditions could improve the results obtained when applying the GAN to images using low lighting, as shown in Figure~\ref{figure3}.
More perspectives could be generated from the same pallet block, or conversely new pallet blocks (i.e., a new chipwood surface structure) could be generated while retaining the same perspective.
Lastly, further experiments using other GAN architectures should be conducted, as a methodological comparison was beyond the scope of this work.

\section{Acknowledgments}
This work is part of the project “Silicon Economy Logistics Ecosystem” which is funded by the German Federal Ministry of Transport and Digital Infrastructure.

\bibliography{app_of_synth.bib}
\end{document}